\relax
\documentclass[letterpaper]{article} 
\usepackage{aaai21}  
\usepackage{times}  
\usepackage{helvet} 
\usepackage{courier}  
\usepackage[hyphens]{url}  
\usepackage{graphicx} 
\usepackage{natbib}  
\usepackage{caption} 
\usepackage{amsmath,amssymb,amsfonts}
\usepackage[ruled,vlined]{algorithm2e}
\usepackage{url}
\usepackage{color}
\usepackage[normalem]{ulem}
\usepackage[switch]{lineno}

\title{Designing Interpretable Approximations to Deep Reinforcement Learning}
\author{
 Nathan Dahlin,
 Krishna Chaitanya Kalagarla,
 Nikhil Naik,
 Rahul Jain, 
 Pierluigi Nuzzo 
 }
 \affiliations{

     University of Southern California\\

     3740 McClintock Avenue, 
     Los Angeles, CA 90089-2560\\
 }

\begin{document}

\maketitle

\begin{abstract}
In an ever expanding set of research and application areas, deep neural networks (DNNs) set the bar for algorithm performance. However, depending upon additional constraints such as processing power and execution time limits, or requirements such as verifiable safety guarantees, it may not be feasible to actually use such high-performing DNNs in practice. Many techniques have been developed in recent years to \emph{compress} or \emph{distill} complex DNNs into smaller, faster or more understandable models and controllers. This work seeks to 
identify reduced models that not only preserve a desired performance level, but also, for example, succinctly explain the latent knowledge represented by a DNN. 
 We illustrate the effectiveness of the proposed approach on the evaluation of decision tree variants and kernel machines in the context of benchmark reinforcement learning tasks. 
\end{abstract}

\section{Introduction}
\noindent 
Deep neural network (DNN)-driven algorithms now stand as the state of the art in a variety of domains, from perceptual tasks such as computer vision, speech and language processing to, more recently, control tasks such as robotics \cite{mirzadeh2019improved}, \cite{bastani2018verifiable}. Nevertheless, there is often reason to avoid direct use of DNNs. For example, the training from scratch or hyperparameter tuning of such networks can be prohibitively expensive or time consuming \cite{schmitt2018kickstarting}. For some applications, the size or complexity of such DNNs precludes their use in real time, or employment in edge devices with limited processing resources \cite{chen2017learning}, \cite{mirzadeh2019improved}. In other areas such as flight control or self driving cars, DNNs are sidelined (at least for mass deployment) by their opaqueness or lack of decision making interpretability \cite{bastani2017interpretability}, \cite{DBLP:conf/aies/HindWCCDMRV19}. 

In light of such issues, a vast body of work has focused in recent years  on developing simpler or more structured controllers which retain desirable properties of a given DNN based controller \cite{bastani2018verifiable}. Additionally, multiple studies in this area, e.g., \cite{bastani2018verifiable}, \cite{ba2014deep} attribute the efficacy of DNNs across such a broad range of problems not to an inherently richer representative capacity over other architectures, or even over shallower neural networks, but rather to the many regularization techniques which currently facilitate DNN training. Therefore, while it may not always be clear how to precisely obtain alternative controllers with performance similar to DNNs, it is theoretically possible, and therefore well-motivated, to do so. 

A related challenge is how to select an alternative controller, given multiple objectives. For example, given a reference DNN, one may wish to design an alternative controller with fewer parameters and comparable performance that is also easier to understand. In this work, we propose a collection of metrics constituting a framework for evaluating how well surrogate controller \textit{distilled} from a target DNN ``match'' the original, as well as how to make comparisons amongst competing alternatives. We focus on reinforcement learning (RL) tasks, and in particular on decision tree and kernel machine alternatives to DNNs trained via the DQN algorithm \cite{mnih2013playing}. We study standard hard decision trees based on thresholding of input attributes at inner nodes as well as ``soft'' decision trees as described in \cite{frosst2017distilling} and in later sections of this work. 

Beyond average reward, when discretization of the environment state space is tractable, we consider policy accuracy percentage, number of parameters, and normalized root mean square (NRMS) error between what we term the \textit{empirical value functions} (EVFs) associated with each controller (including the reference DQN). To obtain the EVFs, we 
use sample trajectories starting from a collection of points spanning a given RL environment's state space, since the alternative controllers we consider do not directly provide such an estimate. 
Based on these metrics, we assess collections of hard and soft trees, as well as kernel machines, 
In summary, \textit{our contributions} can be summarized as follows:
\begin{itemize}
    \item We propose a collection of metrics including a novel, empirical value function based RMS distance metric for evaluating the quality of distilled hard and soft decision tree controllers;
    \item We train hard and soft decision trees of varying depth, along with kernel machines of with varying kernel function and misclassification penalty parameters, and examine the impact of this parameter on our proposed distillation quality metrics. Particularly in the kernel machine case, few such studies exist in the literature. 
\end{itemize}

\section{Background}\label{section:background}
\subsection{Related Work}
Related literature can be broken into two primary areas:  knowledge distillation and model evaluation metrics. 

\subsubsection{Knowledge distillation} 
In essence, distillation is the transfer of behavior or learned knowledge for a given problem or task from one model or controller to another. Briefly, this is related to, but distinct from \emph{model compression}, which seeks to quantize, code or otherwise process reference network weights in a way that leads to a reduced complexity model of the same structure \cite{polino2018model}. 

In \cite{bucilua2006model}, modestly sized models are trained on ``pseudo-data'' generated by large ensembles of base level classifiers. This teacher-student paradigm is central to the distillation literature. Shallow neural networks are trained in \cite{ba2014deep} 
to achieve comparable performance to state-of-the art DNNs with the same number of parameters when the shallow networks are trained to mimic the DNNs instead of learning directly from the original labeled training data. Concentrating on edge devices such as smart phones,  \cite{mirzadeh2019improved} demonstrate a student of fixed size or complexity will perform poorly if the teacher is too large, and propose a ``teaching-assistant'' facilitated process involving multiple distillation steps between the original teacher and target student. 

\cite{hinton2015distilling} argues that training a student neural network using a weighted combination of the correct labels and the output soft labels (class probabilities) generated by a teacher network helps transfer knowledge to the student regarding relative likelihoods of different classes, therefore improving student performance. This approach is extended in \cite{chen2017learning} to more complex, multi-class object tasks by incorporating training loss functions accounting for class imbalance as well as ``hints'' from intermediate teacher layers, amongst other techniques. Introducing \emph{soft decision trees} (SDTs) which essentially feature a single layer perceptron at each inner node, \cite{frosst2017distilling} demonstrated that the target student need not be a smaller neural network. Training these SDTs on data generated by an expert DNN improves performance over training directly on the labeled data. See Section \ref{subsection:SDT} for a more detailed overview of this approach.

Distillation has also been extended from the realm of supervised learning to RL. Using a technique termed \textit{policy distillation}, \cite{rusu2015policy} shows the policy of an RL agent can be extracted to train a smaller, more efficient network to perform at expert level. \cite{bastani2018verifiable} augment the \textsc{Dagger} algorithm \cite{ross2011reduction} by making use of the expert network's Q-function (see Section 2.1) to extract a series of policies, the best of which is selected based upon a cross-validation procedure. \cite{liu2018toward} extend mimic learning to RL settings via Linear Model U-Trees (LMUTs), a variant of the U-Tree representation of a Q function \cite{mccallum1996learning}, placing linear models at each leaf node. 
Finally, nearest to our experimental work, \cite{coppens2019distilling} studied how the SDTs described in \cite{frosst2017distilling} can be used to explain the behavior of expert DNNs in an RL setting. 

\subsubsection{Evaluation metrics} The collection of controller evaluation metrics presented here is most closely related to \cite{andrews1995survey}, which proposes four metrics to measure the quality of rules extracted from a neural network \cite{dancey2007logistic}: accuracy, fidelity, consistency and comprehensibility. For a classifier $\hat{c}$, accuracy is defined in \cite{andrews1995survey} as 
$$P(\hat{c}(X) = C),$$
where $C$ represents a previously unseen problem instance. Fidelity, the extent to which the classifer $\hat{c}$ decisions correspond to the original neural network $(nn)$ is defined as the probability 
$$P(\hat{c}(X) = nn(X)).$$ Consistency refers to the stability of the extracted policy across multiple $nn$ training sessions, and comprehensibility is the number of rules extracted from the network, along with the number of antecedants or conditions per rule. 


Our EVF based NRMS calculation is most closely related to the examination of mean absolute error (MAE) and root mean square error (RMSE) between reference DNN and LMUT Q function representations in \cite{liu2018toward}. We instead examine the distance between empirical estimates for the \emph{value} functions corresponding to each type of controller, as this captures in some sense the policy actually followed by each controller, given the system dynamics. 

\subsection{Imitation Learning}
We use the methodology of \textit{imitation learning} (IL) 
\cite{ross2011reduction} to train our simpler and more interpretable students (decision trees and kernel machines) to imitate the complex trained DQN (expert). IL trains a classifier or regressor to predict the behavior of an expert based on a dataset of observation (state) and action trajectories produced by the expert. While extensions of this basic approach which account for discrepancies between the distributions of states visited by the expert and student are available e.g., \cite{ross2011reduction}, we employ the basic IL approach here and leave exploration of more advanced techniques to future research.



\subsection{Soft Decision Trees}\label{subsection:SDT}
While DNNs perform well in automatically learning policies for control problems, this efficacy is often offset by a lack of clarity as to how specific actions are selected. 
Beyond the first or last few layers, it is usually hard to explain the functional role of a given node as layers repeatedly aggregate latent features in order to form subsequent representations. 

Decision trees offer one alternative, often more legible decision making paradigm. Decision tree selected actions can be traced back through sequences of decisions based directly on input data. Apart from traditional decision trees, which split left and right at the root and inner nodes based upon thresholding of a single input feature, we examine \emph{soft decision trees} (SDTs) as described in \cite{frosst2017distilling}, and refer to the former as \emph{hard decision trees} (HDTs). In SDTs, each inner node $i$ learns a filter $\textbf{w}_i$ and bias $b_i$, in order to output a probability of taking the right branch
$$p_i(\textbf{x}) = \sigma(\beta(\textbf{x}^T\textbf{w}_i + b_i)).$$
The `inverse temperature' parameter $\beta$, is introduced in order to avoid very soft decisions, and $\sigma$ is any suitable activation function. Each leaf node $\ell$ learns a distribution over the possible output selections
$$Q^{\ell}_k = \frac{\exp(\phi^{\ell}_k)}{\sum_{k'}\exp(\phi^{\ell}_{k'})}.$$
See \cite{frosst2017distilling} for more details of the training process for this decision tree variant. 

Unlike standard decision trees, this architecture allows for decisions to be made at each inner node based on aggregated input characteristics, rather than splitting based upon value ranges of input characteristics. 

Empirically it has been shown \cite{frosst2017distilling} that on supervised learning tasks such as MNIST digit recognition, SDTs trained on the outputs of a neural net expert exhibit better generalization than trees trained on data directly, though they do not match the performance of the expert. However, it is often easier to explain the output decisions of the SDTs than those of the neural network, as SDTs learn to make hierarchical decisions. 

Our SDT implementation is based on the Github repository ``Hierarchical mixture of Bigots" \cite{SDT_code} associated with publication \cite{frosst2017distilling}. We use the Python sklearn DecisionTreeClassifier class to implement HDTs \cite{scikit-learn}. 


\subsection{Kernel Machines}

As a second more interpretable alternative to DNN based controllers, we consider kernel machines (KMs). Kernel machines make predictions on unlabeled inputs based upon weighted comparisons with a collection of labeled training data points or \emph{support vectors}. Thus, kernel machines provide interpretability in the sense that the decision made for a given input sample can be explained in terms of the contribution of each support vector, i.e. how similar the input is to each support vector. The similarity between new inputs and support vectors is calculated based on a function known as the \emph{kernel}. For example, in the case of binary classification, label prediction for a new input $\textbf{x}$ takes the form \cite{kernel_methods}
\begin{equation}\label{KM_decision}\hat{y}(\textbf{x}) = \text{sign}\sum_{i=1}^n\alpha_iy_ik(\textbf{x}_i,\textbf{x}) + b,\end{equation}

where $\textbf{x}_i$, $y_i\in\{-1,+1\}$, $\alpha_i\in\mathbb{R}_+\,\forall\,i$ and $b$ are the training examples, example labels, learned weights and offset, respectively and $k$ is a real valued kernel function. Any function which corresponds to an inner product in some inner product space constitutes a valid kernel. In this work we focus on the commonly used \emph{Gaussian} or \emph{radial basis} kernel, given by : 

\begin{equation}
    k(\textbf{x}_i,\textbf{x}) = \exp(-\gamma\|\textbf{x}_i-\textbf{x}\|^2),
\end{equation}

where $\gamma>0$ is an input parameter. 

The weights and offset are determined by solving a multiobjecive optimization problem on the training dataset which maximizes the margin between input classes while penalizing misclassifications with regularization parameter $C$. We use the sklearn SVC class to implement our kernal machines \cite{scikit-learn}. 


\section{Controller Characterization Metrics}\label{section:metrics}
Given a reference expert controller and a collection of distilled models, it is desirable to identify which of the models represents the ``best'' approximation to the expert. 
Much recent work on distillation techniques and their application focuses on controller performance \cite{bastani2018verifiable}, \cite{coppens2019distilling}, \cite{frosst2017distilling}. 
Still, controller performance alone does not indicate the extent to which the distilled controller ``matches'' or explains the target DNN policy, particularly in cases where fitted models have lower performance than the expert \cite{coppens2019distilling}. Aside from performance, \emph{fidelity}, or the extent to which fitted models match the predictions of the expert is another natural evaluation metric for distilled models \cite{liu2018toward}.  

In Section \ref{section:casestudy} and Section \ref{section:carracing_casestudy} we evaluate surrogate  student models distilled from a reference DQN trained DNN on the basis of controller performance as well as fidelity metrics. 

In cases where discretization of the full state space is tractable, we examine two fidelity metrics. 
First, taking the view of both the reference DNN and students as function approximators for the true value function of a given environment, we examine the normalized RMS error between the \emph{empirical value functions} associated with each model. 

\begin{figure}[h]
    \centering
    \includegraphics[width=8cm]{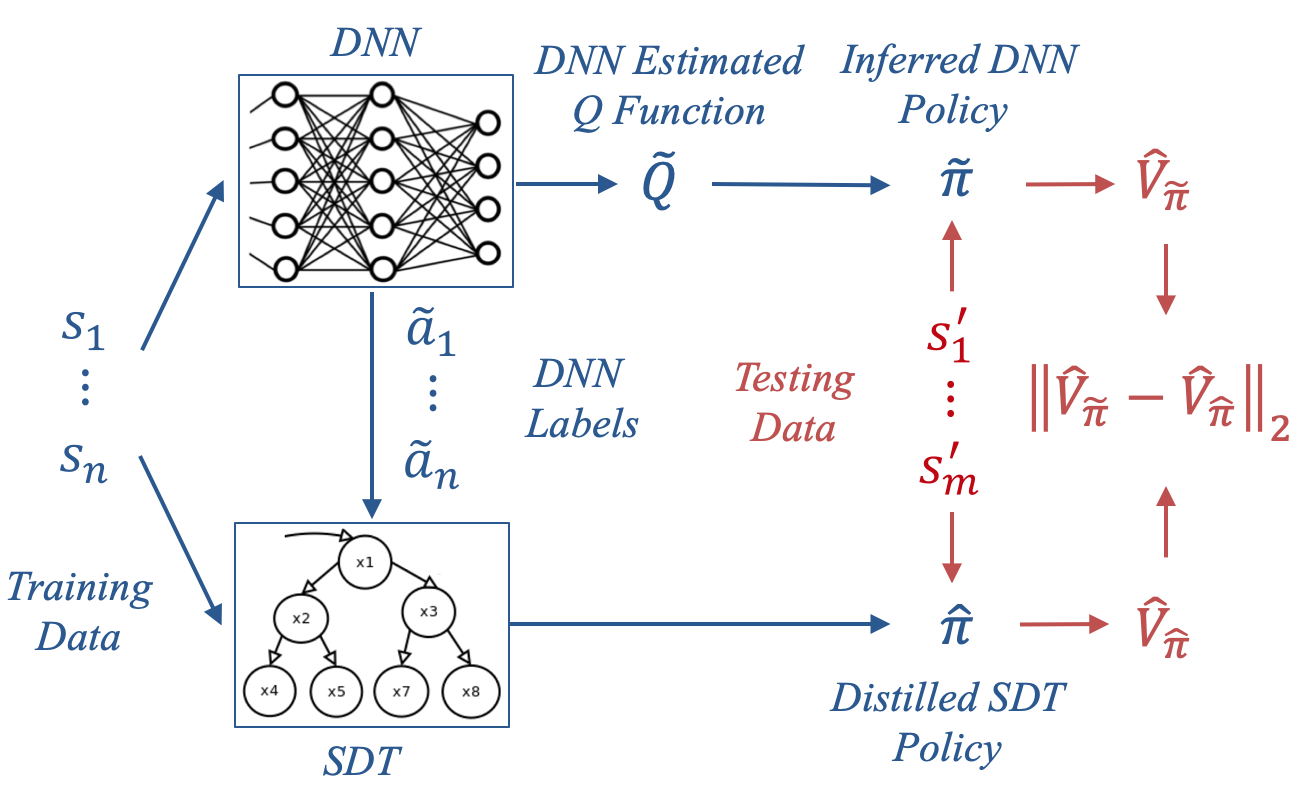}
    \caption{Generation and L2 norm comparison of empirical value functions $\hat{V}_{\tilde{\pi}}$ and $\hat{V}_{\hat{\pi}}$.}
    \label{fig:empvalfcndiagram}
\end{figure}

Figure \ref{fig:empvalfcndiagram} illustrates the generation and comparison of these empirical value functions when the student is an SDT. The reference DNN can be represented by the state-action value or $Q$ function,  $\widetilde{Q}$. Given distillation training observation set (states) $s_{\text{train}} = \{s_1,\dots,s_n\}$, with $n\in \mathbb{Z}$, the DNN outputs corresponding labels (actions) $\widetilde{a} = \{\widetilde{a}_1,\dots,\widetilde{a}_n\}$. Together, labeled data $(s,\widetilde{a})$ is used to train a students as detailed in Section \ref{section:background}, and we infer policies $\widetilde{\pi}$ and $\widehat{\pi}$ corresponding to the DNN and student, respectively. 
Starting from randomly seeded starting states $s_{\text{test}} = \{s'_1,\dots,s'_m\}$ with $m\in\mathbb{Z}$, trajectories for both the DNN and SDT are generated in order to obtain empirical estimates of the value functions $\widehat{V}_{\widetilde{\pi}}$ and $\widehat{V}_{\widehat{\pi}}$ associated with $\widetilde{\pi}$ and $\widehat{\pi}$, respectively. We then take the empirical L2 norm of the difference between $\widehat{V}_{\widetilde{\pi}}$ and $\widehat{V}_{\widehat{\pi}}$, evaluated at $s_{\text{test}}$ and normalized by the maximum absolute value of $\widehat{V}_{\widehat{\pi}}$:
\begin{equation}\nonumber\text{RMSE}(\widehat{V}_{\widetilde{\pi}},\widehat{V}_{\widehat{\pi}},s_{\text{test}}) = \sqrt{\frac{1}{m}\sum_{s'\in s_{\text{test}}}\left(\widehat{V}_{\widetilde{\pi}}(s') - \widehat{V}_{\widehat{\pi}}(s')\right)^2}\end{equation}
\begin{equation}\label{eqn:rms_nrms}\text{NRMSE}(\widehat{V}_{\widetilde{\pi}},\widehat{V}_{\widehat{\pi}},s_{\text{test}}) = \frac{\text{RMSE}(\widehat{V}_{\widetilde{\pi}},\widehat{V}_{\widehat{\pi}},s_{\text{test}})}{\max_s\,|\widehat{V}_{\widehat{\pi}}(s)|}.
\end{equation}

We focus here on value function approximation, rather than Q-value function approximation, as the value function reflects the actions that are actually taken under a given policy, and therefore captures in some sense the closed loop behavior of the controllers under comparison. 

Secondly, we consider is policy accuracy, or percent 0-1 loss, which we later report as the percentage
$$\text{\%ACC}(\widetilde{\pi},\widehat{\pi},s_{\text{grid}}) = \frac{|\{s\in s_{\text{grid}}\,:\,\widetilde{\pi}(s) = \widehat{\pi}(s)\}|}{|\{s\in s_{\text{grid}}\}|},$$
where $s_{\text{grid}}$ is a discretization of $S$. 

In cases where the environment state space cannot be efficiently discretized, we resort to policy matching accuracy on a validation dataset. 

Aside from performance and fidelity, we also consider the complexity of candidate controllers in terms of the number of included trainable parameters. 

\section{Case Studies}\label{section:casestudy}

\subsection{MountainCar-v0}
In this section, we detail the MountainCar-v0 environment case study. A similar analysis of the CartPole-v0 environment can be found in Appendix A. 



\subsubsection{Problem description}
The mountain car problem is a well-known benchmark problem in reinforcement learning, prevalent in the literature since the 1990s \cite{moore1990efficient}. 
The problem includes an under powered mountain car trying to reach a hilltop, starting out from a valley. As the car lacks sufficient power to drive directly uphill to the goal, the optimal strategy is to reverse up the hill on the opposite side and use the acquired momentum in addition to the engine to reach the goal. 

In this problem, the state space is continuous with two states -- the position of the car along the $x$--axis, contained in the interval $[-1.2,0.6]$, and the velocity of the car, contained in the interval $[-0.07,0.07]$. Car position is initialized randomly in range $[-0.6,-0.4]$, with velocity set to 0. Three actions are allowed -- the car can either choose to go left, go right or do nothing. The task is declared successfully completed if the car reaches the goal state within 200 time steps. For every time step that the goal is not reached, the car receives a reward of $-1$ point. The episode is terminated as a failure if the time limit is exceeded and the car has not reached its goal.

\subsubsection{Expert DQN}
For MountainCar-v0, our expert neural net consisted of two hidden layers with 24 and 48 units with ReLu activation, and three units with linear activation to account for the three possible actions. Discount factor $\gamma$ was set to 0.99 for training, and this architecture was trained using the DQN algorithm on 400 episodes of MountainCar-v0. The output reference model has a total of 1419 trainable parameters. The DQN learned the optimal control policy successfully. We tested the DQN controller over 100 episodes, and observed that the controller succeeded in achieving the goal in every episode. 
The DQN completed the control task over all the 100 episodes with a mean cumulative reward and 95\% confidence interval of 
-154.05$\pm 0.79$ units. 

Both soft and hard trees were trained on a set of labeled data generated by the expert DQN over 1500 MountainCar-v0 episodes. The data was preprocessed such that the final distillation data contained equal numbers of state/actions pairs for each of the three available actions. In all, the distillation training set contained 
167,085 data points. 

\subsubsection{Hard and soft decision trees} Soft decision trees of depths 2 through 9 were trained using the technique discussed in Section \ref{subsection:SDT}. Hard trees of depths 2 through 9 were trained on the DQN generated labeled data.

Figures \ref{fig:NRMSL2Error} through \ref{fig:sdt_num_params} display the results of the application of the metrics introduced in Section \ref{section:metrics} to the sets of hard and soft trees, and reference DQN. Starting with our fidelity metrics, Figure \ref{fig:NRMSL2Error} compares the NRMS values as calculated in (\ref{eqn:rms_nrms}) 
for each tree type and depth. For the purposes of this plot, the state space was discretized to 20 steps in each dimension, giving 400 test trajectories overall for each controller. As can be seen, the L2 error does not vary considerably with the tree depth for trees of either type, though the error increases slightly with depth for hard trees.

Figure \ref{fig:sdt_perc_0_1} shows the policy accuracy percentage for each of the distilled controllers. In this experiment, the state space was discretized more finely 
into 100 steps for each dimension. For the tree depths tested, depth 5 gives the optimal accuracy for SDTs, as deeper, more complex trees appear to overfit to the DQN labeled data. On the other hand, in terms of percentage policy accuracy, HDTs increase monotonically with depth over the range tested. 

Turning to performance, Figure \ref{fig:sdt_meanconfidenceintervals} plots the empirical mean and 95\% confidence intervals across 100 test trajectories for each controller. The DQN outperforms all SDTs in terms of mean reward, though the 95\% confidence bounds of the depth 5 SDT overlap with those of the DQN. HDTs of depth 7 or greater either slightly outperform or essentially match the DQN. 

Finally, Figure \ref{fig:sdt_num_params} compares the complexity of each distilled tree and the reference DQN in terms of tunable parameters. For each SDT, this number represents the learned weights and biases of each inner node, along with the probability distributions of each leaf node. For each HDT, this number is twice the number of inner nodes, as each inner node splits the data based upon an input attribute and threshold. For tree depth larger than 7, the SDTs actually become more complex than the DQN in the sense just described. The highest performing, depth 5 SDT is specified by 190 parameters, meaning this tree requires about 13\% of the parameters needed to specify the DQN. The HDTs of depth 7, 8 and 9, which outperform this depth 5 SDT require 330, 446, and 570 parameters, respectively.

\begin{figure}[h]
    \centering
    \includegraphics[width=8cm]{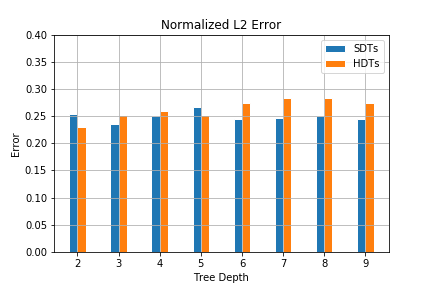}
    \caption{Normalized RMS L2 Error for HDT/SDTs depth 2-9 (statespace discretized into 20 steps per dimension).}
    \label{fig:NRMSL2Error}
\end{figure}
\begin{figure}[h]
    \centering
    \includegraphics[width=8cm]{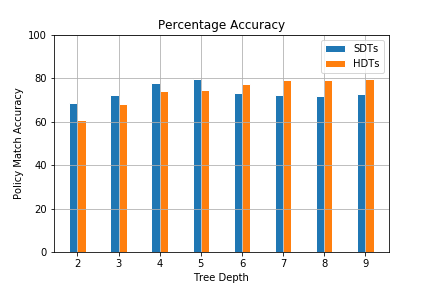}
    \caption{Percentage Policy Accuracy for HDT/SDTs depth 2-9 (statespace discretized into 100 steps per dimension).}
    \label{fig:sdt_perc_0_1}
\end{figure}
\begin{figure}[h]
    \centering
    \includegraphics[width=8cm]{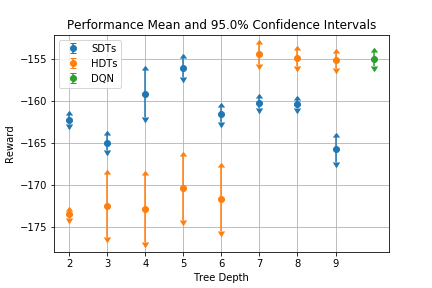}
    \caption{Performance evaluation for HDT/SDTs and reference DQN for 100 episodes.}
    \label{fig:sdt_meanconfidenceintervals}
\end{figure}
\begin{figure}[h]
    \centering
    \includegraphics[width=7cm]{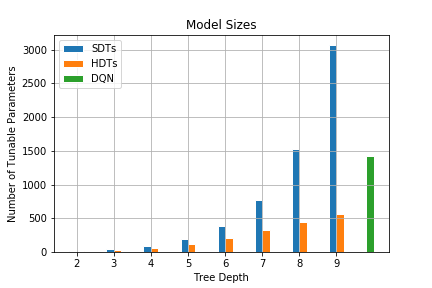}
    \caption{Number of parameters for HDT/SDTs and reference DQN.}
    \label{fig:sdt_num_params}
\end{figure}
\subsubsection{Discussion} Overall, we identify three primary takeaways from our case study. First, larger tree depth does not always result in improved performance or fidelity. Second, given our choice of decision tree architecture and training methodology, as well as RL environment, it seems that the percentage accuracy is a better predictor of performance than the proposed L2 error metric. Third, soft decision trees with a fraction of the tunable parameters of the original DNN can achieve similar performance in the MountainCar-v0 task. 

\subsection{CarRacing-v0 Case Study}\label{section:carracing_casestudy}

In this section, we detail the CarRacing-v0 environment case study. 

\subsubsection{Problem Description}

\begin{figure}[h]
    \centering
    \includegraphics[width=7cm]{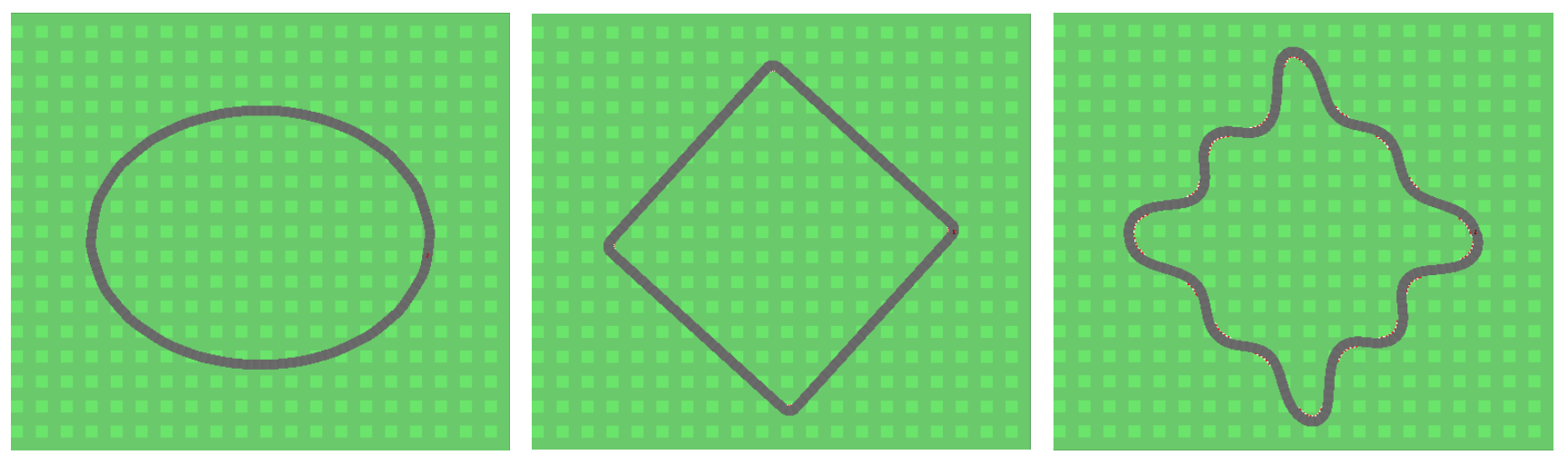}
    \caption{CarRacing-v0 track set: elliptical, square and zigzag.}
    \label{fig:track_set}
\end{figure}

CarRacing-v0 is a top-down, two dimensional racing environment regarded as the easiest OpenAI gym control task to learn from pixels. The task in this environment is to complete a single lap of a race track, the shape of which can be configured. In particular, we trained and evaluated all controllers on one or more of three tracks displayed in Figure \ref{fig:track_set} : elliptical, square and zigzag.   

The state space for CarRacing-v0 consists of $96\times 96$ pixel image frames. The environment provides RGB image values, but for this study we converted images to gray scale for all controllers. Actions consist of turning (-1 for left, +1 for right, 0 for straight), acceleration (zero or fixed value) and braking (zero or fixed value) decisions, giving 12 possible actions in total. The track loop is formed from a series of connected tiles. The reward is -0.1 for each step plus $1000/N$, where $N$ is the total number of track tiles. For a given track, the car position is initialized to the same point in each episode, with velocity set to 0. Episodes terminate when either all tiles are visited, or a specified number of consecutive negative rewards accumulates.

\subsubsection{Expert DQNs}

For each of the three tracks mentioned above, we trained a corresponding DNN. We also trained a DNN on the full set three tracks, with training episodes repeatedly cycling through the tracks. Each DNN was comprised of two convolutional/maxpooling layers, followed by flattening and and two fully connected layers. The first convolutional layer includes 8 channels of $7\times 7$ filters, the second convolutional layer includes 16 channels $3\times 3$ filters with both layers using ReLu activatoin, and both followed by $2\times 2$ maxpooling. The flattened layer output is a 1024 entry vector. The first fully connected layer has 128 units with ReLu activation, and the second has 12 units with linear activation. 

Each DNN was trained using the DQN algorithm. In each case targeting a single track, DQN learned a successful policy, i.e., a policy navigating the car through a full loop with score of at least 850. For the three-track case, the DQN learned a policy which was successful across all tracks. 

Hard decision trees and Gaussian kernel machines (KMs) were trained on labeled data in each case. Soft decision trees were not successful in completing the task, and are not considered in this section. For the individual track experiments, the training data consisted of 10 episodes of frame/action trajectories, giving 2310 samples for the elliptical track, 2500 samples for the square track, and 3320 samples for the zigzag track. For the three-track experiment, the training data consisted of 4 episodes of frame/action trajectories for each track, giving 4068 samples. In all experiments, we randomly partitioned the trajectory data into a 90/10 split of testing and validation datasets. 

\subsubsection{Hard Decision Trees and Kernel Machines}

Hard decision trees of depths 2 through 20 were trained and tested, although due to poor results for smaller trees we focus here on 
depths 8 through 20. Gaussian kernel machines were trained and tested, with all combinations of kernel parameter $\gamma\in[0.1,1,10]$ and regularization parameter $C\in[0.1,1,10]$.

\begin{figure}[h]
    \centering
    \includegraphics[width=7cm]{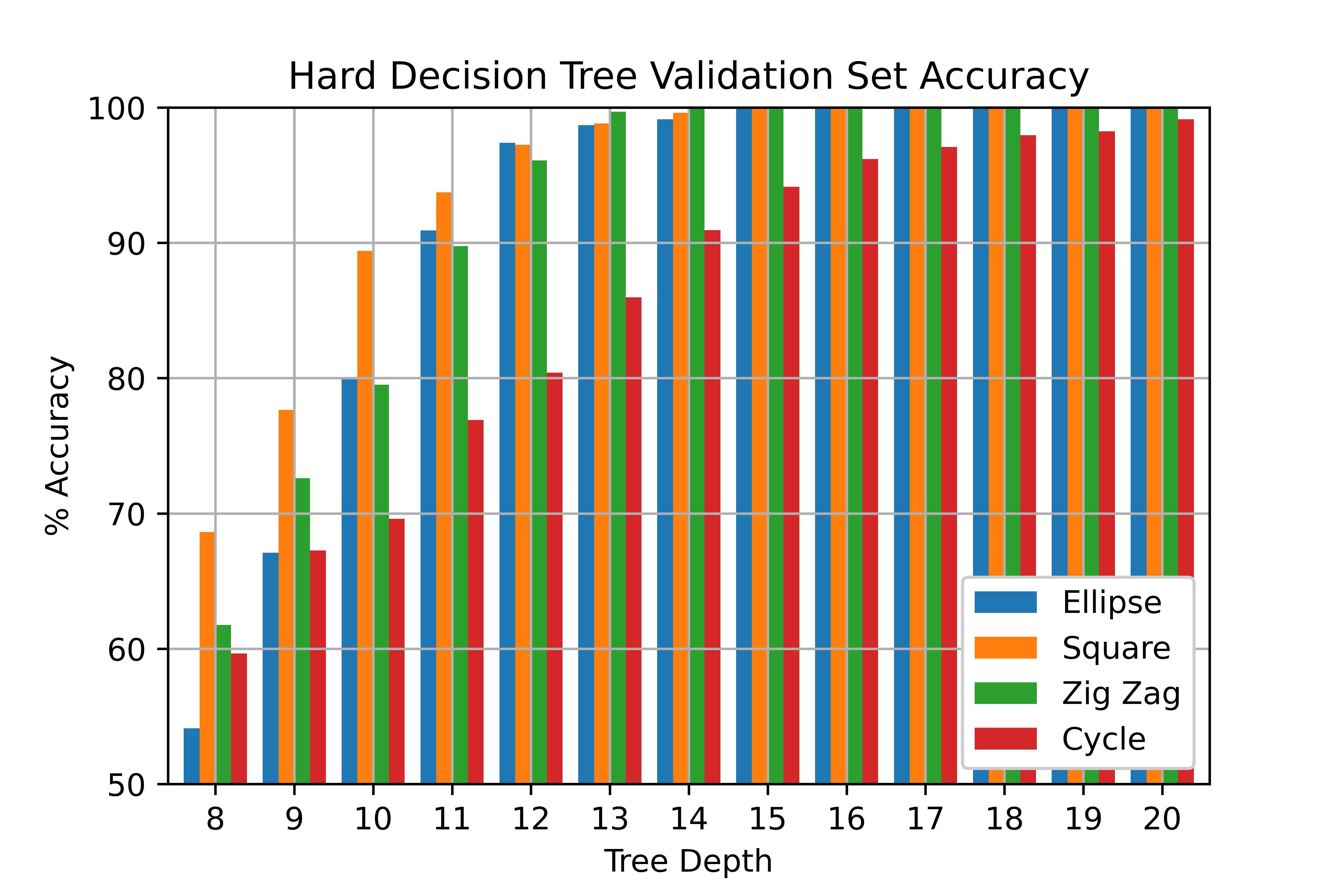}
    \caption{Car Racing HDT validation set accuracy.}
    \label{fig:car_racing_HDT_acc}
\end{figure}

\begin{figure}[h]
    \centering
    \includegraphics[width=7cm]{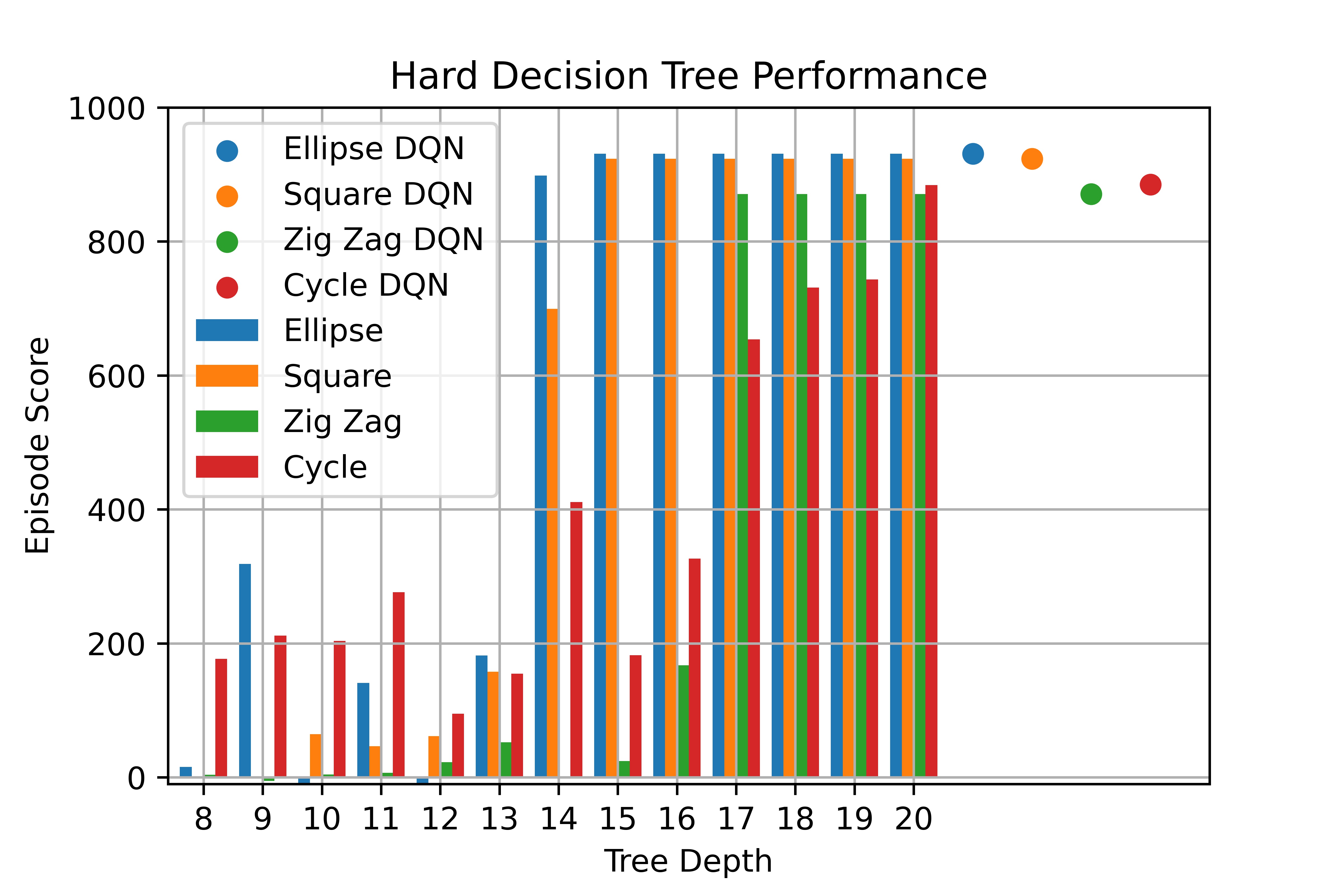}
    \caption{Car Racing HDT performance.}
    \label{fig:car_racing_HDT_perf}
\end{figure}

Figure \ref{fig:car_racing_HDT_acc} illustrates the validation set percentage accuracy against DQN actions across a range of tree depths. For each individual track, decision trees of depth 16 or higher achieve near perfect accuracy. For the three track set, deeper improve monotonically, reaching above 99\% accuracy by depth 20. In Figure \ref{fig:car_racing_HDT_perf}, it can be seen that decision trees of depth 16 and above match the DQN performance, while for the three track set, depth 20 is required to match DQN performance. 

\begin{figure}[h]
    \centering
    \includegraphics[width=7cm]{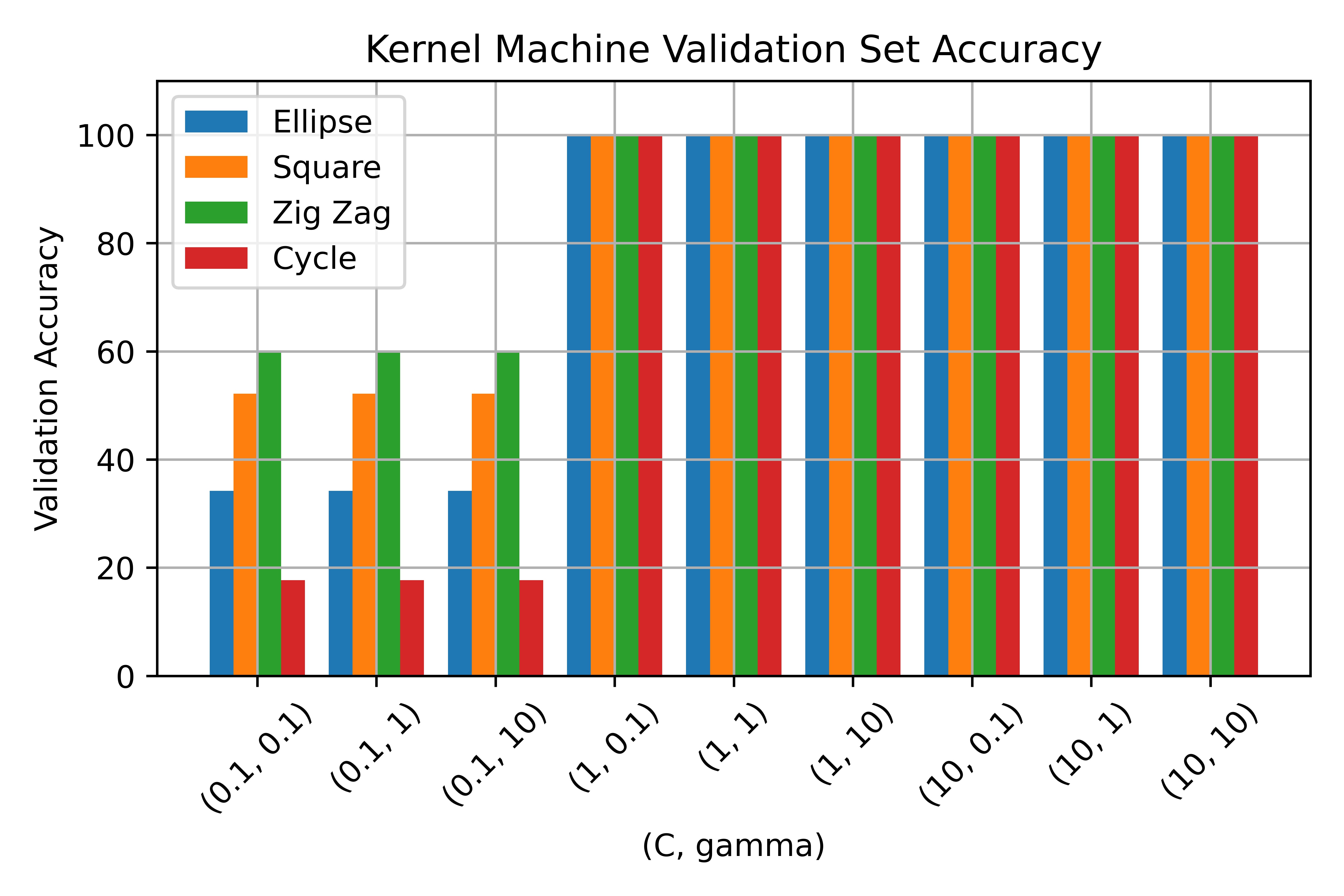}
    \caption{Car Racing KM validation set accuracy.}
    \label{fig:car_racing_KM_acc}
\end{figure}

\begin{figure}[h]
    \centering
    \includegraphics[width=7cm]{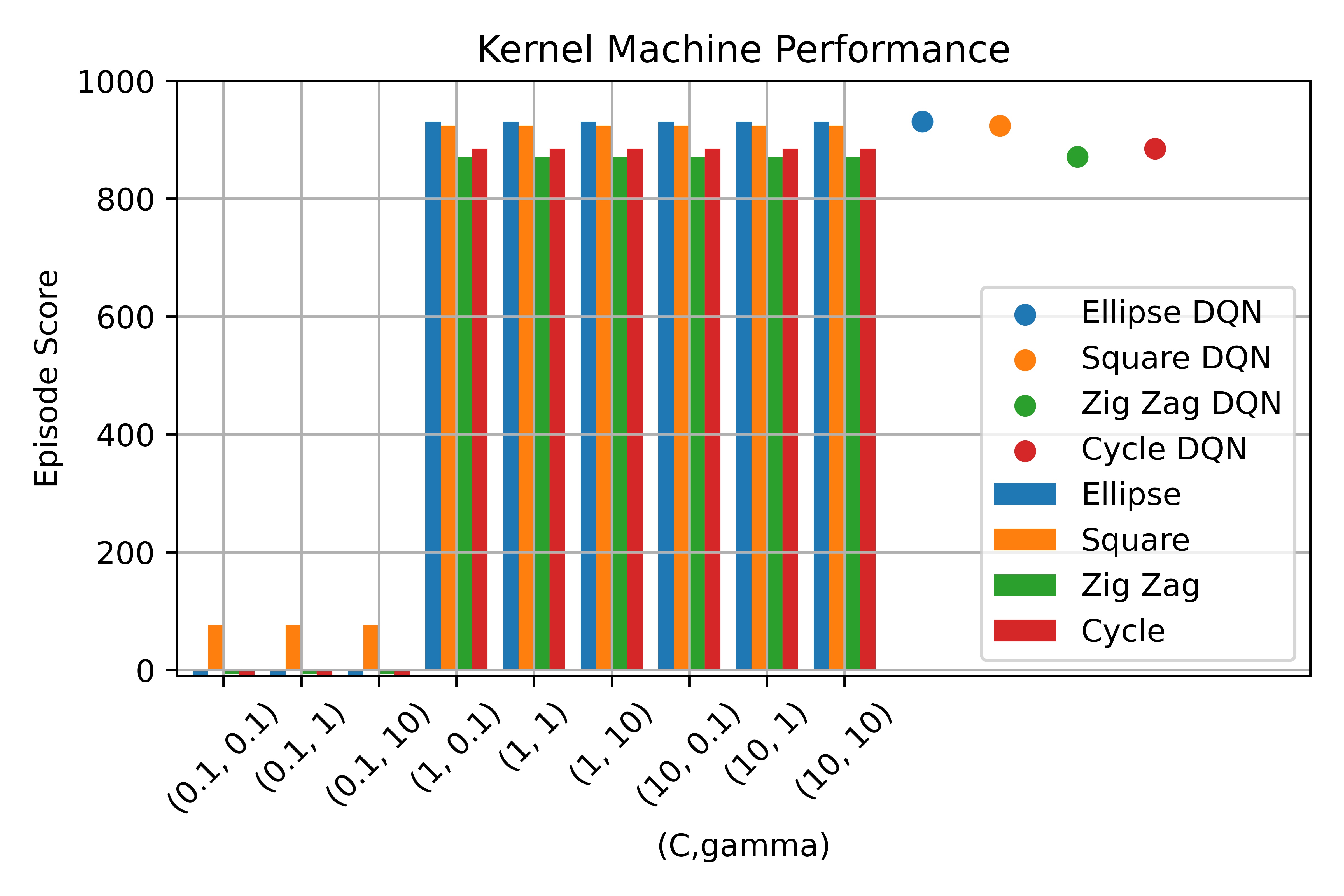}
    \caption{Car Racing KM performance.}
    \label{fig:car_racing_KM_perf}
\end{figure}

Similarly, figures \ref{fig:car_racing_KM_acc} and \ref{fig:car_racing_KM_perf} plot the validation set accuracy and performance results for the Gaussian kernel machines. In all experiments, including the three track experiment, kernel machines using regularization parameter $C=1$ or $C=10$ achieve perfect accuracy, and match the DQN performance, regardless of the value of $\gamma$.

\subsubsection{Discussion}

Overall, both HDTs and KMs are capable of matching DQN performance, and accurately mimicking the DQN policy on the respective validation datasets. While in the case of HDTs the results change considerably with tree depth, the KM results are relatively insensitive to parameter choices. Another notable characteristic of the KMs is that the best performing machines are the simplest in terms of number of support vectors used in the online decision function. Taking the three track experiment as an example, KMs with $C=0.1$ use 99\% of the training set input points, while those with $C=1$ and $C=10$ use 70\% and 28\% of the input points, respectively. In contrast, decision trees generally require more depth, i.e. more branches and leaf nodes in order to attain DQN level performance. 

Again, for lower values of regularization parameter $C$, less penalty is incurred for misclassification, so that more training samples can be identified as support vectors than in the cases of higher $C$ values. However, together figures 9 and 10 show that high accuracy is crucial to the performance of the kernel machine controllers, as validation set accuracy of up to roughly 60\% still results in an essentially unusable controller in terms of performance. 

\section{Conclusion and Future Work}

We introduced an evaluation framework and metrics to assess the quality of distilled decision tree controllers in RL settings against a variety of design objectives. Specifically, we consider normalized root mean square error between empirical value functions, policy match accuracy, mean performance with confidence bounds, and number of tunable parameters to guide designers and researchers toward assessing tradeoffs between performance, fidelity, and complexity across a variety of potential surrogate controllers. 

Future directions for our work include 
incorporating more sophisticated imitation learning algorithms such as \textsc{Dagger} \cite{ross2011reduction} into the training of our alternative architectures, and developing metrics which focus more on the ``closed-loop'' behavior of controllers in a given environment. 

\bibliography{ijcai21}
\bibliographystyle{aaai}

\appendix
\section{CartPole-v0 Case Study}

In this section, we detail the CartPole-v0 environment case study referenced in the primary submission. 



\subsection{Problem description}
The cart pole problem is another benchmark problem in reinforcement learning \cite{barto1983neuronlike}. 
The specific version of the problem implemented in the CartPole-v0 OpenAI-Gym environment and studied here includes a pole attached to a cart via an un-actuated joint. The cart moves along a frictionless track, attempting to maintain the pole's initial upright position. 

The state space is continuous, with four elements: cart position, cart velocity, pole angle (measured from vertical with positive angles to the right) and pole velocity (measured at the tip). The cart position is contained within the interval $[-2.4,2.4]$, while the pole angle is contained in the interval $[-41.2^{\circ},41.2^{\circ}]$. Both cart and pole angle velocities may take on any real value. Each state attribute is initialized to a uniformly distributed value in range $[-0.05,0.05]$. Two actions are available, push left or push right. Episodes terminate when either the absolute value of the pole angle exceeds $12^{\circ}$, the cart position reaches $\pm 2.4$, or the episode lasts 200 time steps. The reward is 1 for each time step taken.

\subsection{Expert DQN}
For CartPole-V0, our neural network included 4 dense layers, giving 38,531 trainable parameters overall. We used a baseline DQN framework from the OpenAI Gym baselines package \cite{baselines} for implementing the DQN algorithm. We tested the DQN controller over 100 episodes, and observed that the controller succeeded in obtaining the maximum possible reward of 200 in each episode.

Both soft and hard trees were trained on a set of labeled data generated by the expert DQN over 750 CartPole-v0 episodes. The data was preprocessed such that the final distillation data contained equal numbers of state/actions pairs for each of the two available actions. In all, the distillation training set contained 
148,452 data points. 

\subsection{Hard and soft decision trees} Soft decision trees (SDTs) of depths 2 through 9 were trained using the technique discussed in \cite{frosst2017distilling}. Hard decision trees (HDTs) of depths 2 through 9 were trained on the DQN generated labeled data using the sklearn Python package \cite{scikit-learn}. 

\begin{figure}[t]
    \centering
    \includegraphics[width=8cm]{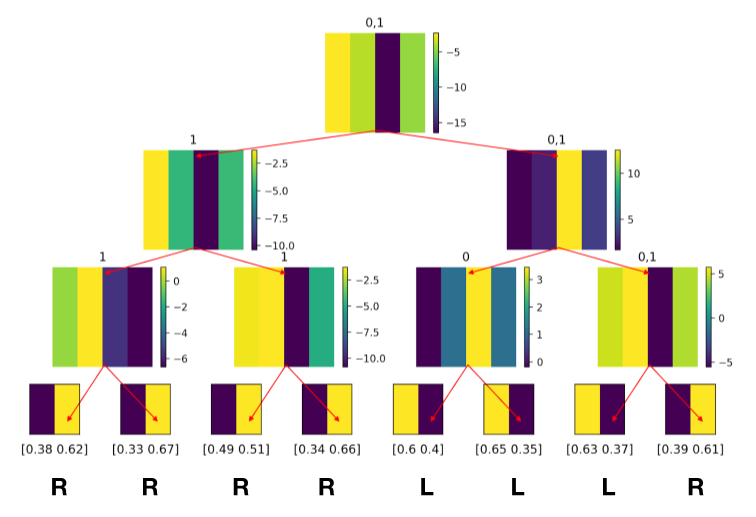}
    \caption{Soft Decision Tree for Cart Pole}
    \label{fig:sdt_mountaincar}
\end{figure}

Figure \ref{fig:sdt_mountaincar} shows a trained SDT of depth 3. The colored squares in the inner layer nodes display the trained weights $\textbf{w}_i$. At each inner node, the four panels give the weight applied to each input feature, in the order described above, from left to right. Thus, for example, one may observe that the decision made at the root node depends most heavily on the pole angle. The two-color panels at the leaf nodes represent the learned probability distribution over potential actions, with the letter corresponding to highest probability action given for each leaf node in the bottom row. For this particular tree, there is a clear highest probability action corresponding to each leaf node, aside from the third from the left. 

Figures \ref{fig:NRMSL2Error} through \ref{fig:sdt_num_params} display the results of the application of the metrics introduced in Section 3 of the primary submission to the sets of hard and soft trees, and reference DQN. Starting with our fidelity metrics, Figure \ref{fig:NRMSL2Error} compares the NRMS values as calculated according to
\begin{equation}\nonumber\text{RMSE}(\widehat{V}_{\widetilde{\pi}},\widehat{V}_{\widehat{\pi}},s_{\text{test}}) = \sqrt{\frac{1}{m}\sum_{s'\in s_{\text{test}}}\left(\widehat{V}_{\widetilde{\pi}}(s') - \widehat{V}_{\widehat{\pi}}(s')\right)^2}\end{equation}
\begin{equation}\nonumber\text{NRMSE}(\widehat{V}_{\widetilde{\pi}},\widehat{V}_{\widehat{\pi}},s_{\text{test}}) = \frac{\text{RMSE}(\widehat{V}_{\widetilde{\pi}},\widehat{V}_{\widehat{\pi}},s_{\text{test}})}{\max_s\,|\widehat{V}_{\widehat{\pi}}(s)|}.
\end{equation}
for each tree type and depth. For the purposes of this plot, the state space was discretized to 5 steps in each dimension, giving 625 test points overall. The cart position was discretized across $[-2.4,2.4]$, while the pole angle was discretized across $[-12^{\circ},12^{\circ}]$, i.e., the nonterminating values. Both cart and pole velocities were limited to range $[-0.05,0.05].$  As can be seen, for both types of trees the error generally decreases with tree depth up to depth 8.

Figure \ref{fig:sdt_perc_0_1} shows the policy accuracy percentage for each of the distilled controller. In this experiment, the state space was discretized more finely, as each data point requires a call to the controller's prediction function, rather than a complete episode trajectory. In particular, the range of each input state feature was discretized into 10 steps, over the ranges mentioned in the preceding paragraph. For the tree depths tested, depth 9 gives the optimal accuracy for SDTs. On the other hand, percentage policy accuracy remains essentially constant for HDTs.

Turning to performance, Figure \ref{fig:sdt_meanconfidenceintervals} plots the empirical mean and 95\% confidence intervals across 100 test trajectories for each controller. Trees of both types are successful in matching the performance of the DQN. All but depth 3 HDTs achieve the maximum reward of 200, along with SDTs of depth 2 through 5, as well as 7, within confidence bounds. 

Finally, Figure \ref{fig:sdt_num_params} compares the complexity of each distilled tree and the reference DQN in terms of tunable parameters. The DQN parameter count far exceeds any tree tested, and thus is omitted. For each SDT, this number represents the learned weights and biases of each inner node, along with the probability distributions of each leaf node. For each HDT, this number is twice the number of inner nodes, as each inner node splits the data based upon an input attribute and threshold. The smallest SDT to achieve a mean reward of 200, depth 3, is specified by 52 parameters, while the smallest HDT to achieve this mean is specified by just 6 parameters. 

\begin{figure}[h]
    \centering
    \includegraphics[width=8cm]{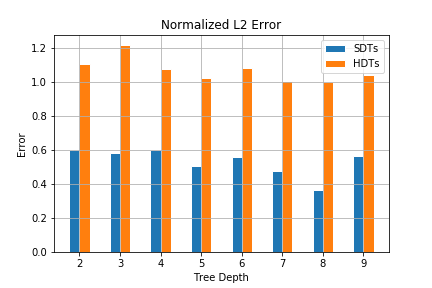}
    \caption{Normalized RMS L2 Error for HDT/SDTs depth 2-9 (statespace discretized into 20 steps per dimension).}
    \label{fig:NRMSL2Error}
\end{figure}
\begin{figure}[h]
    \centering
    \includegraphics[width=8cm]{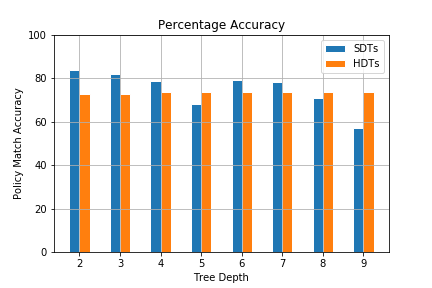}
    \caption{Percentage Policy Accuracy for HDT/SDTs depth 2-9 (statespace discretized into 10 steps per dimension).}
    \label{fig:sdt_perc_0_1}
\end{figure}
\begin{figure}[h]
    \centering
    \includegraphics[width=8cm]{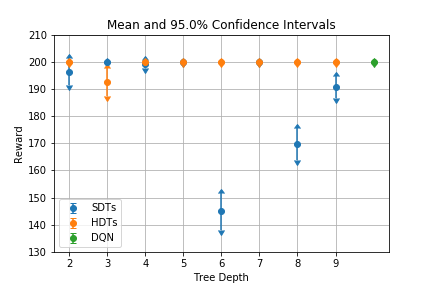}
    \caption{Performance evaluation for HDT/SDTs and reference DQN for 100 episodes.}
    \label{fig:sdt_meanconfidenceintervals}
\end{figure}
\begin{figure}[h]
    \centering
    \includegraphics[width=7cm]{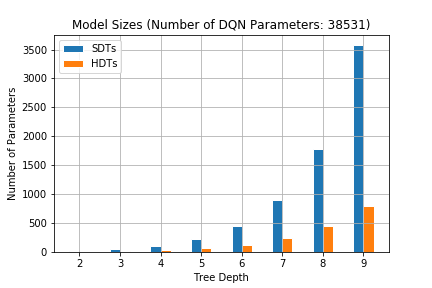}
    \caption{Number of parameters for HDT/SDTs and reference DQN.}
    \label{fig:sdt_num_params}
\end{figure}
\subsection{Discussion} For the CartPole-v0 task, the clearest correspondence between metrics comes in the case of HDTs, where essentially uniform accuracy across tree depth aligns with basically uniform performance across tree depth. Within confidence bounds, SDTs which exceed the HDT level of accuracy also match the DQN performance level, aside from depth 6. It is important again note that each episode is initialized uniformly in the range $[-0.05,0.05]$ for each state attribute, meaning that the range of states examined for both L2 error and percentage accuracy likely exceeds the range of states encountered by the reference DQN, as well as other high performing controller. This may explain the lack of obvious correspondence between performance and either of these metrics in SDTs, and HDTs in the case of L2 error, as many states tested may not be encountered often in actual trajectories. Finally, both soft and hard decision trees with a fraction of the tunable parameters of the original DNN can achieve similar performance in the CartPole-v0 task. 




\end{document}